\pdfoutput=1

\documentclass[11pt]{article}

\usepackage[preprint]{acl}

\usepackage{times}
\usepackage{latexsym}
\usepackage{booktabs, multirow}
\usepackage[T1]{fontenc}  
\usepackage[utf8]{inputenc}  
\usepackage{microtype}  
\usepackage{inconsolata}  
\usepackage{graphicx}  
\usepackage{xspace}

\usepackage{graphicx}
\usepackage{adjustbox}
\usepackage{pgf}
\usepackage{colortbl}

\definecolor{lightgray}{gray}{0.9}

\newcommand*{\escape}[1]{\texttt{\textbackslash#1}}

\everymath=\expandafter{\the\everymath\displaystyle}

\usepackage[font={small},labelfont={bf}]{caption}
\usepackage{authblk} 

\title{Beyond Demographics: Fine-tuning Large Language Models to Predict Individuals' Subjective Text Perceptions}

\author[1]{\textbf{Matthias Orlikowski}}
\author[2]{\textbf{Jiaxin Pei}}
\author[4]{\textbf{Paul R\"ottger}}
\author[1]{\textbf{Philipp Cimiano}}
\author[3]{\textbf{David Jurgens}}
\author[4]{\textbf{Dirk Hovy}}
\affil[1]{Bielefeld University}
\affil[2]{Stanford University}
\affil[3]{University of Michigan}
\affil[4]{Computing Sciences Department, Bocconi University, Milan, Italy}

\newcommand{\Sref}[1]{\S\ref{#1}}
\newcommand{\sref}[1]{\S\ref{#1}}

\newcommand\dataset{\textsc{DeMo}\xspace}

\begin{document}
\maketitle

\begin{abstract}
People naturally vary in their annotations for subjective questions and some of this variation is thought to be due to the person's sociodemographic characteristics. LLMs have also been used to label data, but recent work has shown that models perform poorly when prompted with sociodemographic attributes, suggesting limited inherent sociodemographic knowledge. Here, we ask whether LLMs can be trained to be accurate sociodemographic models of annotator variation. Using a curated dataset of five tasks with standardized sociodemographics, we show that models do improve in sociodemographic prompting when trained \textit{but} that this performance gain is largely due to models learning annotator-specific behaviour rather than sociodemographic patterns. Across all tasks, our results suggest that models learn little meaningful connection between sociodemographics and annotation, raising doubts about the current use of LLMs for simulating sociodemographic variation and behaviour. 
\end{abstract}

\section{Introduction}

Most NLP models require labelled data to learn. Yet, the humans labelling that data may not agree what is the correct label. These annotator disagreements stem from multiple causes, such as genuine mistakes, adversarial behaviour, or even individual preferences \citep{sandri2023don}. This variance in labelling behaviour has long been recognized and multiple models have been developed to distinguish some types of disagreements, particularly those due to error \citep{hovy-etal-2013-learning,passonneau-carpenter-2014-benefits}. Recent work has focused on modelling the regularity in label variation due to individual \citep[e.g.,][]{deng-etal-2023-annotate} and group-based preferences \citep[e.g.,][]{davani2024d3code}. For example, members of one social group may regularly rate a piece of text as more or less offensive than others. When such labelling behaviour is regular, a large language model (LLM) could be prompted to take on sociodemographic characteristics to generate how a person with such characteristics would answer the question \citep{beck-etal-2024-sensitivity}.

\begin{figure}[t]
  \centering
   \includegraphics[width=0.45\textwidth]{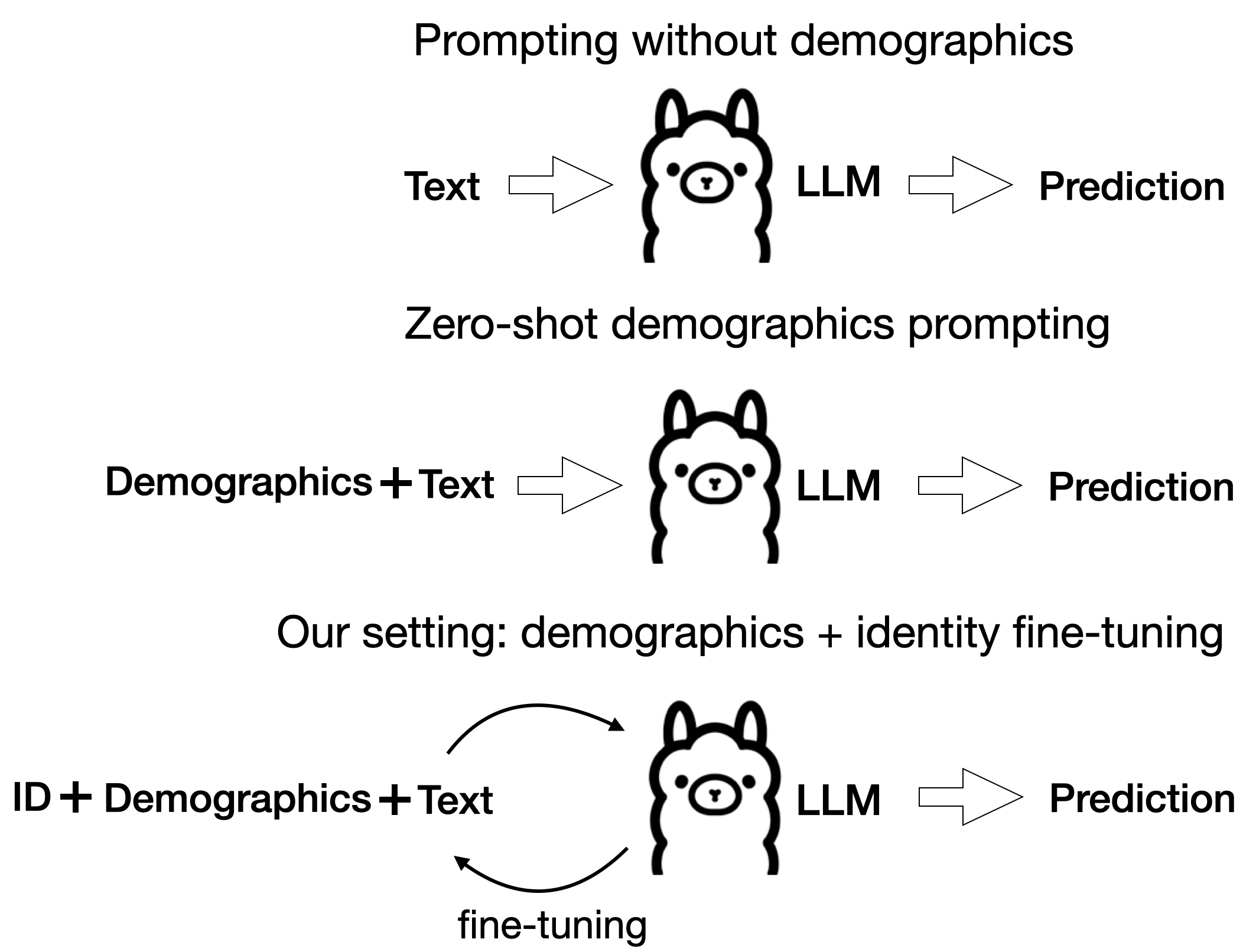}
  \caption{Unlike existing works that majorly rely on zero-shot demographic prompting, we explore whether LLMs can be \emph{trained} to predict individuals' subjective text perceptions.}
 \label{fig:reg_demographics_datasets}
 \end{figure}

Such approaches of sociodemographic prompting require that an LLM can effectively take the perspective of the person or group in the prompt. When LLMs are used to generate synthetically-labelled data \citep{grunde2023designing} or as evaluators \citep{dong2024can}, this approach provides a scalable way to include meaningful variation by annotators, particularly those for less common sociodemographic identities \citep{simmons2023large}. However, multiple works have raised issues with the accuracy of this approach when LLMs are used in zero-shot settings \citep[e.g.,][]{beck-etal-2024-sensitivity,hu-collier-2024-quantifying,sun2023aligning}. While the root cause of this low zero-shot performance is likely multifaceted, given the potential benefits of LLMs as annotator models, \emph{we test whether LLMs can be trained as sociodemographic models of annotators}, which was not assessed before.

To effectively model individual annotators with LLMs, we introduce a new approach that combines persona prompting with annotator modelling. Instead of fixing annotator identity and attributes as part of a specialized architecture, we incorporate this information by adapting input formats from persona prompts. Using this formatted input, we fine-tune decoder-only LLMs with prediction heads as used in reward models for LLM alignment \citep[e.g.,][]{liu_skywork-reward_2024}. For training, we curated the \dataset{} dataset by unifying five existing datasets for subjective classification tasks (intimacy, offensiveness, politeness, safety, sentiment) that include annotator IDs and sociodemographic attributes (age, gender, race, and education). 

Our study answers the following four research questions.
\textbf{RQ1}: Can LLMs learn to model \emph{given} annotators better based on sociodemographics or their identity (ID)? LLMs improve over baselines when incorporating sociodemographics, but we find that LLMs are much more accurate at modelling specific annotators' behaviours.
\textbf{RQ2}: Can LLMs generalize to \emph{new} annotators?
No, we find that neither sociodemographic attributes nor IDs improve performance over a text-only baseline, suggesting LLMs do not learn generalizable patterns.
\textbf{RQ3}: If LLMs can use sociodemographic attributes to better model given annotators (RQ1) but do not generalize (RQ2), what information do LLMs learn when improving from sociodemographics?
We find that sociodemographic-tuned models primarily improve for annotators with unique attributes, where attributes effectively act as an ID, suggesting LLMs are learning little about the connection between sociodemographics and annotation behaviour.
\textbf{RQ4}: Does modelling annotator identity improve how models predict label distributions when annotators disagree?
Beyond improvements in predicting individual ratings, we show that models using annotator identity better reflect cases of disagreement between annotators than a text-only baseline.
All data and code realted to our experiments are available at {\small{\url{https://github.com/morlikowski/beyond-demographics}}}.

\section{Related Work}
Within research on the role of annotator characteristics in annotation,  we connect work in sociodemographic prompting with annotator modelling.

\paragraph{Annotator Characteristics in Annotation.}
\label{sec:relatedwork-annotation-behaviour}
Training NLP and AI models relies on human annotations to adjust their parameters to align with human knowledge and preferences. Unlike games and mathematics, where there is always a ground truth, many NLP annotation tasks are inherently subjective. They are affected by annotators' attributes and individual preferences. Existing studies have explored how different attributes affect annotators' behaviours on tasks such as sentiment analysis \cite{diaz2018addressing}, preference modelling \cite{kirk2024PRISM}, ideology classification \cite{shen-rose-2021-sounds} and hate speech or toxicity detection \citep{larimore-etal-2021-reconsidering, kumar_designing_2021, sap-etal-2022-annotators}. Effects seem to be strongest when annotated content and attributes align (e.g., LGBTQ identities in relation to homophobic content, \citealt{goyal_is_2022}), but are also found across different tasks for more general samples \cite{pei-jurgens-2023-annotator}. However, similar to us, some works do not find relevant associations with annotator background \cite{biester-etal-2022-analyzing}. Consequently, recent studies explore differences within demographic groups \cite{davani2024d3code}.

\paragraph{Annotator Modelling}
\label{sec:relatedwork-annotator-modelling}
Annotator  investigates supervised models that predict the annotations of individual annotators on specific inputs. These works are motivated by wider research on annotator subjectivity that questions the assumption of a single ground truth in annotation \cite{ovesdotter-alm-2011-subjective,uma-2021-survey,plank-2022-problem,fleisig-etal-2024-perspectivist,frenda_perspectivist_2024}. Our work builds on studies that model annotators in subjective tasks \cite{davani-etal-2022-dealing, weerasooriya-etal-2023-disagreement, vitsakis-etal-2023-ilab, wang-plank-2023-actor}. Many annotator models use a unique identifier (ID) per annotator, often represented as a learnt embedding, frequently in combination with information derived from annotation statistics \cite{heinisch-etal-2023-architectural, oluyemi-etal-2024-corpus, mokhberian-etal-2024-capturing}. However, for unseen annotators, \newcite{deng-etal-2023-annotate} show that models with annotator embeddings (ID and annotation patterns) do not beat a content-only baseline. We also evaluate on an annotator-based split (see \sref{sec:rq4}). However, as we focus on the availability of sociodemographic metadata, only one task, Sentiment (see \sref{sec:demo_dataset}), overlaps with datasets used in their study. \newcite{anand-etal-2024-dont} learn from individual annotations and evaluate the confidence of predictions. In particular, they find improvements in high-disagreement instances, similar to our analysis in \sref{sec:analysis-disagreement}.

Closely related to our work, some annotator models include sociodemographic information on annotators (e.g., \citealt{wan_everyones_2023}). \citet{orlikowski-etal-2023-ecological} and \citet{fleisig-etal-2023-majority} find conflicting results on the usefulness of sociodemographics relative to annotator identity which we discuss in relation to our findings (see \sref{sec:discussion}). Other works find improvements from demographics over a content-only baseline but do not compare to using only annotator IDs \cite{gordon-etal-2022-jury, jaggi_accurate_2024}. In concurrent work, \citet{jiang-etal-2024-examining} also present a study on fine-tuning LLMs with annotator information, as part of a dataset description and analysis. In contrast to our study, they do so in the context of only a single dataset with few training instances (600) and also do not compare against using the ID. Their results, similar to our findings, indicate that sociodemographics are less influential, showing greater importance for attitudes directly related to their task \cite{jiang-etal-2024-examining}.

\paragraph{Sociodemographic Prompting and Simulation.}
\label{sec:relatedwork-prompting}
Sociodemographic prompting is part of a broader interest in using LLMs to simulate human responses in surveys or experiments. Within the social sciences, simulations focusing on simple actors and macro patterns from interactions are an established method \cite{epstein_growing_1996}. In contrast, LLMs enable human surrogates for new settings such as surveys or experiments, which several studies have started to explore \cite{aher_simulate_2023, dillion_llms_2023, kozlowski_simulating_2024}. While some studies report successful applications \cite{argyle_out_2023,horton2023large,manning2024automated}, others discuss downsides of using LLMs to simulate individuals based on background descriptions, such as caricature and misportrayal of social groups \cite{cheng-etal-2023-compost, wang_misportray_2024}.
Among these, our work builds on investigations into simulating annotators by prompting LLMs based on sociodemographic profiles \cite{beck-etal-2024-sensitivity, hu-collier-2024-quantifying}. In concurrent work, \citet{gao_scylla_2024} find that prompted LLMs do not align well with human outcome distributions in a behavioural experiment but that LLMs fine-tuned on relevant examples do, similar to our findings (see \sref{sec:rq3-results}).

\begin{table*}[t]
\centering
\resizebox{\textwidth}{!}{%
\begin{tabular}{ c m{3.5cm} c c c r r r r}
 \hline
Task & Labels & Reference & Data Type & Instances & Raters & Labels per Instance & Total Labels \\
\hline
\rowcolor{gray!10}
Intimacy & Not Intimate to Very Intimate (1-5) & \citealt{pei-etal-2023-semeval} & Tweet & 1,993 & 261 & 48 & 12,516 \\

Offensiveness & Not Offensive to Very Offensive (1-5) & \citealt{pei-jurgens-2023-annotator} & Reddit comment & 1,500 & 262 & 50 & 13,036 \\
\rowcolor{gray!10}
Politeness & Not Polite to Very Polite (1-5) & \citealt{pei-jurgens-2023-annotator} & Email & 3,718 & 506 & 50 & 25,042 \\
Safety & Yes, Unsure, No & \citealt{aroyo2024dices} & Conversation & 350 & 104 & 350 & 36,050 \\
\rowcolor{gray!10}
Sentiment & Very Negative to Very Positive (1-5) & \citealt{diaz2018addressing} & Tweet & 14,071 & 1,481 & 41 & 60,654 \\
\hline
\end{tabular}%
}
\caption{The datasets used in \dataset}
\label{tab:data_description}
\end{table*}

\section{\dataset{}}
\label{sec:demo_dataset}
We curate a collection of five published datasets containing annotations and sociodemographic annotator information. These datasets focus on subjective text perceptions like sentiment and offensiveness and represent a diverse range of tasks and sociodemographics. We identify the largest intersection of sociodemographic attributes across the datasets. All five datasets contain information on gender, age, race, and education, so we select these four attributes for our experiments. To provide more comparable analyses, we normalize the sociodemographic attributes of the five tasks into a consistent and unified set of attributes. Appendix \ref{sec:preprocessing-attributes} details this normalization process and the final attributes used in our datasets. 

The data collectively contains 21,632 texts annotated by 2,614 annotators resulting in 147,297 annotations total. Table \ref{tab:data_description} shows the statistics of each dataset. and how we pre-processed attributes across datasets. Below we briefly introduce each task and the original dataset on which it is based.

\paragraph{Intimacy} Intimacy reflects the perceived closeness of messages in interpersonal communications, and we use the English subset of the \textsc{MinT} dataset \citep{pei-etal-2023-semeval} which contains 1,993 tweets annotated by 261 annotators. Each tweet is annotated by 7 annotators with an intimacy score from 1 (``Not intimate at all'') to 5 means (``Very intimate'').

\paragraph{Offensiveness} The perception of offensiveness \cite[i.e., language that might cause displeasure, anger, or hurt feelings ][]{chinivar2023online} is subjective and depends on individual attributes like gender and race \citep{jacobi2014perceptions}.  We use the offensiveness subset of the \textsc{Popquorn} dataset \citep{pei-jurgens-2023-annotator}. It includes 13,036 annotations from 1,500 annotators for 1,500 Reddit comments and nuanced demographic information of the annotators. 

\paragraph{Politeness} Politeness refers to ``linguistic behavior which is perceived to be
appropriate to the social constraints of the ongoing interaction'' \citep{watts2003politeness}. It is one of the most fundamental concepts of interpersonal communication. We use the 25,042 politeness annotations from 506 annotators in the  \textsc{Popquorn} dataset \citep{pei-jurgens-2023-annotator}.

\paragraph{Safety} focuses on the perceived conversational safety in human-AI interactions, and we use the DICES-350 data \citep{aroyo2024dices}, which contains ratings of conversational safety for degrees of harm using a multifaceted rubric. These data contain 36,050 annotations from 104 annotators when applying the authors' filtering of low-quality annotators.

\paragraph{Sentiment} Sentiment is naturally a subjective construct and individual attributes actively affect people's perception of text sentiment \citep{kumar2020exploring}. We use the sentiment annotations collected by \citet{diaz2018addressing}, which includes 60,654 annotations from 1,481 annotators. 

\section{Experimental Setup}

Here, we describe the different methods and setups for testing LLMs as models for annotators.

\subsection{Weights and Architecture}
\label{sec:weights_architecture}
We fine-tune Llama 3 8B \cite{grattafiori_llama_2024} for our experiments. Llama 3 was among the strongest open-weights models when we started our experiments. We use a standard architecture for learning a prediction head based on a decoder-only transformer, using the implementation for Llama 3 in the HuggingFace Transformers library (\citealt{wolf-etal-2020-transformers}, see Appendix \ref{sec:trainig-details} for implementation details). We use this type of architecture as it is used in current reward models for LLM alignment (e.g., \citealt{liu_skywork-reward_2024}), which is also a task of predicting ratings for text input, similar to the tasks in our experiments. As we fine-tune models with a prediction head and do not rely on instruction-following, we use the Llama 3 base model instead of a post-trained model. In supplementary experiments on a smaller scale, we use Mistral 7B \cite{jiang_mistral_2023} to gauge how results with Llama 3 transfer to other model families, finding minimal differences (details in Appendix \ref{sec:eval_model_families})

\subsection{Data Partitions}

We use two data partition settings for how we split the data into train, validation and test sets to evaluate different aspects of model generalizability. 
The first is the \textit{instance split} where we partition by instance, but annotators might be seen across all three subsets. In this context, an instance means a single text, e.g., a Reddit comment or an email. This setup follows the traditional machine learning setup and allows us to measure whether the LLM can generalize to new instances given an annotator's sociodemographics.
The second is the \textit{annotator split} where annotators are partitioned across train, validation, and test sets. In other words, no annotator in the test set is included in the training or validation sets but the same text may be present in all three subsets. Here, the evaluation measures how well the LLM can simulate a new annotator's decisions based on their sociodemographics. Tables \ref{tab:instance_split}  and \ref{tab:annotator_split} in the Appendix show the statistics of the two data partition settings.

\subsection{Prompt Formats}
\label{sec:prompt-formats}
We fine-tune on inputs which include the instance text and different information about annotators. As we fine-tune models with a prediction head and consequently do not rely on instruction-following (see \ref{sec:weights_architecture}), we use inputs with minimal formatting instead of detailed prompts. Below we detail which annotator information we include and how that information is formatted.

\paragraph{Content-Only}
The baseline setting uses only the textual content without any additional formatting. This format ignores all annotators' attributes.

\paragraph{+Attributes (Content and Sociodemographics)}
In inputs using sociodemographics, we list an annotator's age, gender, race, and education. Attributes in \dataset{} are given as short textual descriptions (e.g., the literal text ``Woman''). We preprocess these descriptions by lowercasing, reformatting age groups (``40-44'' to ``40 to 44 years old''), and adding articles where appropriate (``a woman'' instead of ``woman''). We assume that all possible attribute values are known beforehand, i.e., we do not need to preprocess new values (e.g., an unseen age group) during test time. We format the input text and attribute descriptions based on a minimal template: {\tt Annotator: \{RACE\}, \{AGE\}, \{GENDER\}, \{EDUCATION\}\escape{n} Text: \{TEXT\}}.

\paragraph{+ID (Content and ID)}
This format uses each annotator's unique identifier. As the original ID format varies between tasks in \dataset{}, we reformat IDs to numerical values to ensure uniform input across tasks. The template is {\tt Annotator: unique identifier \{ID\}\escape{n} Text: \{TEXT\}}.

\paragraph{+ID+Attributes (Content, ID and Sociodemographics)}
Lastly, we also test a combined input format. An example looks like this: \texttt{Annotator: unique identifier 72, hispanic/latino, 40 to 44 years old, a woman, a college degree\escape{n} Text: This is an example text}.

\subsection{Baseline System}

As a baseline for our fine-tuning experiments, we run zero-shot sociodemographic prompting experiments similar to \citet{beck-etal-2024-sensitivity} and \citet{hu-collier-2024-quantifying}. Specifically, we prompt Llama 3 Instruct 8B with variants of the "Content Only" and "+Attributes" prompts adapted to a chat prompting template derived from \citet{hu-collier-2024-quantifying}. Here, in contrast to fine-tuning, attributes are described in a conversational format, e.g., \textit{The highest degree or level of school that you have completed is a college degree}. We perform minimal robustness checks using 1) a larger model (Llama 3 Instruct 70B, 4-bit quantized) and 2) a prompt variant that simply lists attributes. We include the best results for Llama 3 Instruct 8B in our fine-tuning results plots. More details and full results are in Appendix \ref{sec:prompting}. 

\section{Experiments}

Can LLMs learn to model sociodemographic variation in annotation (RQ1) and generalize to new annotators (RQ2)? To answer these questions, we evaluate Llama 3 8B \cite{grattafiori_llama_2024} fine-tuned with each of the five prompt formats on the instance split and on the annotator split of \dataset{}. 

\subsection{Training and Evaluation}
We fine-tune models with half-precision weights (bf16) using low-rank adaption (LoRA, \citealt{hu_lora_2021}). We learn LoRA weights for all linear layers except prediction layers and initial token embeddings which are fully fine-tuned. We select the learning rate for each input format and task combination based on the best-performing setting in 10 runs evaluated on the validation set. See Appendix \ref{sec:trainig-details} for full fine-tuning details.

We treat each value of the three-point (Safety) or five-point scales (all others) as a class in an individualized classification task. Individualized means that the model's objective is to predict the annotation that a particular annotator assigned to a specific text. This evaluation setting, often used to evaluate annotator models (see \sref{sec:relatedwork-annotator-modelling}), is intentionally different from standard evaluations in NLP where models are evaluated on a single aggregate rating per text. As each class is equally important in predicting annotators' ratings, we compare models based on macro-average $F_1$. For the main experiments, we run each setting with 30 different random seeds. We report the average score over the 30 runs and compute 95\% confidence intervals using bootstrap sampling.

\subsection{Results: Sociodemographic Modelling}
\label{sec:rq3-results}

\begin{figure}[t]
    \centering
    \input{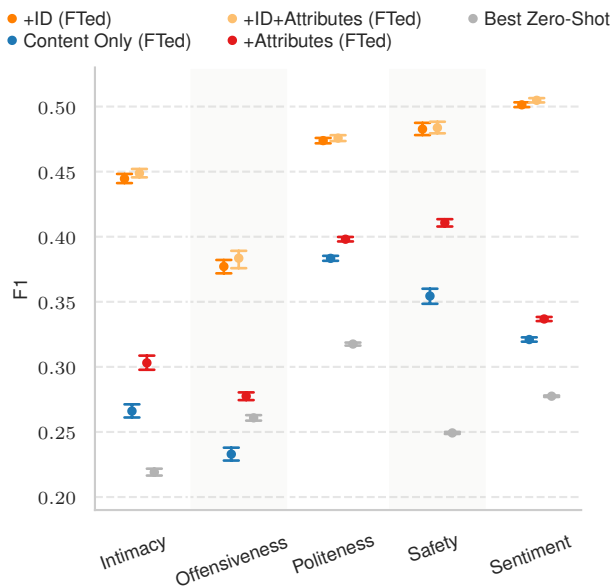}
    \caption{Results on the \emph{instance split} show that training with sociodemographics improves performance over text-only predictions \textbf{but} including a unique annotator ID in the prompt leads to much larger performance gains. Macro-average $F_1$ over three (Safety) or five (all others) classes on each test set. Shows results for Llama 3 8B fine-tuned with different types of input and the best zero-shot result (8B) for each task. Mean score over 30 different seeds with 95\% confidence intervals from bootstrap sampling.}
    \label{fig:rq1-results}
\end{figure}

Including sociodemographic attributes in training significantly outperforms the performance of zero-shot prompting, \textit{initially} suggesting that LLMs can learn to simulate sociodemographic preferences (RQ1), as seen in Figure~\ref{fig:rq1-results} for the instance split. However, when models are prompted with a unique annotator ID, they are even more accurate at predicting the annotator's label. 
For annotators' attributes (red) there is a consistent pattern in contrast to zero-shot LLM behaviour. While in zero-shot, including attributes leads to inconsistent effects, when fine-tuning we see a notable positive shift in the score distribution across all tasks. We analyse this result further in Section \ref{sec:analysis-unique-vs-many}. However, adding the annotator ID (orange) leads to an even higher performance increase. 
When adding both IDs and attributes (light orange), scores are not substantially different from adding only the ID, suggesting that the performance gain from attributes is subsumed by knowledge of who specifically is annotating.

The best zero-shot results per task for Llama 3 Instruct 8B replicate the finding in related work that zero-shot sociodemographic prompting has low performance for individual annotators \citep{beck-etal-2024-sensitivity}. 
Unsurprisingly, even for the content-only baseline (blue), fine-tuning leads to higher scores than zero-shot prompting (grey) for most tasks. A notable exception is the Offensiveness task where the zero-shot performance is slightly above the fine-tuned model. This is due to the task's strong label imbalance where a classifier exclusively trained on the text content can only learn to predict the majority class well, resulting in lower macro-average $F_1$.

\label{sec:rq4}

\subsection{Results: Annotator Generalization}

\begin{figure}
    \centering
    \input{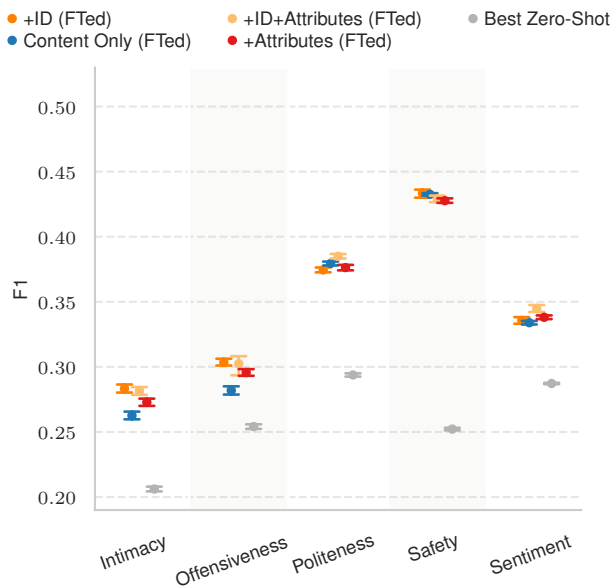}
   \caption{Results on the \emph{annotator split},  where the test sets only include annotators not seen in training, show that training with sociodemographics and/or annotator IDs minimally improve over the text-only baseline. While IDs for unseen annotators is expected to offer little benefit, this result suggests models are not able to generalize from sociodemographics. The plot shows a macro-average $F_1$ is over three (Safety) or five (all others) classes on each test set for Llama 3 8B fine-tuned with different types of input and the best zero-shot result (8B) for each task. Mean score over 30 different seeds with 95\% confidence intervals from bootstrap sampling.}
    \label{fig:rq2-results}
\end{figure}

LLM annotator models do not generalize well to unseen individuals (RQ2), as seen in Figure \ref{fig:rq2-results} the annotator split. While all fine-tuned models perform better than zero-shot results, the performance gains from adding attributes, IDs, or both are negligible compared to the text-only model. When using IDs, no performance gains are expected because the model has not seen these annotators' IDs before and cannot adapt to their idiosyncratic preferences. 
The lack of gains for the sociodemographic attributes suggests that models have, in fact, learned minimal meaningful relationships between text, attributes, and rating combinations. This result is surprising given the results of RQ1 that demonstrated a small but consistent effect from attributes, which suggests that when we have not seen examples from a rater, then their sociodemographic profile should give us at least some information on how they would rate a text. We analyse this result in detail next.

\section{Analyses: What Are LLMs Learning About Sociodemographics?}

The opposite results for sociodemographic prompting in RQ1 and RQ2 suggest that models may not be learning how different attributes influence ratings. Therefore, we perform two additional analyses. First, we assess to what degree are sociodemographic attributes serving as proxies for annotator IDs versus representing meaningful attribute-label relationships (RQ3). Second, we analyse if improvements from including IDs also improve how good models capture cases of disagreement between annotators (RQ4).

\subsection{Sociodemographics as Proxies}
\label{sec:analysis-unique-vs-many}
Given that including IDs improves results much more than attributes on the instance split, we hypothesize that models actually learn to use attribute combinations as a proxy for annotator identity. To test this hypothesis, we compare results for two subsets of annotators: (1) Annotators with \emph{unique} combinations of sociodemographic attributes who have a combination of age, gender, race and education not shared by any other annotator in the test set (denoted \textit{Unique}) and (2) annotators who have a \emph{common} combination of attributes, that is, a sociodemographic profile that is frequently shared by many annotators (denoted \textit{Frequent}). In the former, the attribute combination is effectively a unique identifier for the annotator, while in the latter, the sociodemographics refer to multiple annotators. In the instance split all annotators that are included in the test set are also included in the training set.

For each task, we include all ratings by annotators with a unique profile. For the frequent profiles, we select the top $n$ with $n = 1$ for Sentiment, $n = 3$ for Safety, and $n = 5$ for other tasks. Except for Sentiment, we set $n$ so that the number of annotators is similar for both subsets. For Sentiment, we only use the most frequent profile because it includes more annotators than the sum of unique profiles. More details on profile distributions in Appendix \ref{sec:profiles_distributions}.

To test the hypotheses, we compare the performance gain relative to content-only input for adding each subset of sociodemographic attributes.
For each model configuration and task, we compute new macro-average F1 scores for each subset of annotators across runs.

\begin{figure}
    \centering
    \input{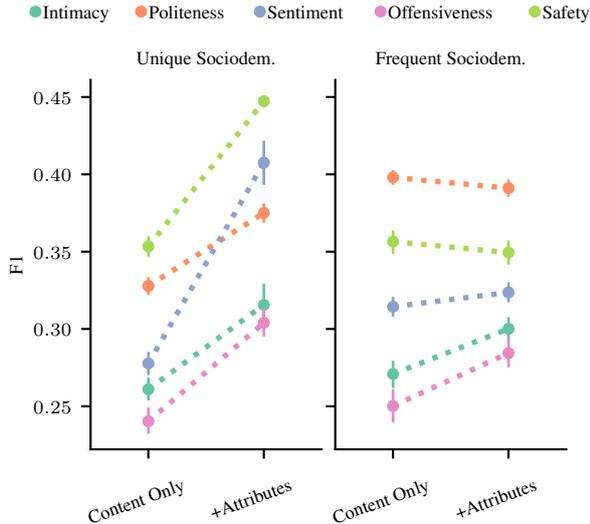}
    \caption{Evaluation scores on ratings by annotators with unique vs. frequent combinations of sociodemographic attributes, corresponding to \textit{Unique} and \textit{Frequent} in the main text. The high improvement for unique sociodemographics compare with the minimal gains for frequent sociodemographics strongly suggests that the LLM is using the attributes as a proxy for annotator ID and is not learning any sociodemographic-label associations. Points show the mean score (macro-average $F_1$) over 30 different seeds for models using only text or text and attributes. Error bars show 95\% confidence intervals from bootstrap sampling.}
    \label{fig:socdem_v2}
\end{figure}

Our results show that the largest gains occur when LLMs are predicting ratings for the annotators in the \textit{Unique} subset (Figure \ref{fig:socdem_v2}), but no consistent or substantial gains for predicting ratings of annotators in the \textit{Frequent} subset. This result confirms our hypothesis that the unique sociodemographics are acting as proxies for identity and, thus, the LLM is not learning any meaningful relationship between attributes and labelling.

\subsection{Modelling Disagreement}
\label{sec:analysis-disagreement}
Beyond getting more accurate at predicting individual ratings, do models improve at capturing specific types of label distributions when incorporating attributes and identifiers (RQ4)? Can we capture when there is disagreement on how to rate an example? Or do models mainly improve on consensually rated content?

To test for which kinds of label distributions the LLM can best model, we group the instances based on their levels of disagreement. We measure disagreement as the entropy of each instance's label distribution: Lower label entropy corresponds to patterns of more agreement and higher label entropy corresponds
to patterns of more disagreement (in the extreme corresponding to uniform ratings). We split instances into two groups using the median value to distinguish lower and higher label entropy. We use the two groups to measure how close models get to predicting the actual label distributions in high and low disagreement scenarios. For each text in the two groups, we measure the distance between the predicted and the actual rating distributions for each model configuration (content only, plus attributes, plus ID).
Following \citet{santurkar_opinionqa_2023}, we compute the distance of the actual rating distributions using Wasserstein distance (earth mover's distance). 

\begin{figure*}
    \centering
    \input{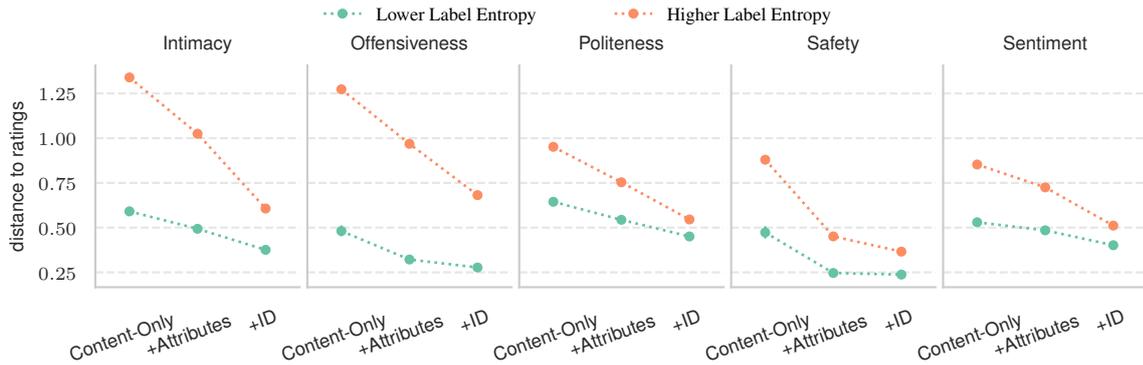}
    \caption{Wasserstein distance to the actual rating distribution (lower is better) on texts in the standard-split test sets, average and 95\% confidence intervals. Lower label entropy corresponds to patterns of agreement and higher label entropy corresponds to patterns of disagreement (uniform ratings or bimodal, diverging ratings). Higher and lower are distinguished based on the median entropy value per test set.}
    \label{fig:disagreement_v2}
\end{figure*}

Models get better at predicting cases of disagreement when including attributes and IDs,  as shown by the scores for higher entropy labels (orange) in Figure \ref{fig:disagreement_v2}. Still, disagreements remain challenging as distances are always higher as for cases when annotators mostly agree. However, distances to the actual rating distribution are smallest on higher disagreement cases when including IDs. With the exception of the Offensiveness task, ID-based models almost model label distributions equally well irrespective of the level of disagreement.
For cases of agreement (teal), there are much smaller differences between model configurations.

\section{Discussion}
\label{sec:discussion}

Based on our results, we can not expect LLMs to model annotators based on their sociodemographics alone, in particular without examples of their individual behaviour. While even the best-performing models still are far from perfect, sociodemographic prompting usually performs worse than using annotator-specific identities. Our results show that it is possible to model a given set of annotators reasonably well from examples, but models do not actually learn how to generalize from seen sociodemographic attribute patterns to new annotators. 
Thus, models only \textit{seemingly} improve from attribute information. As we show, they instead improve for annotators who can be identified by unique attribute combinations. Naturally, this works best if models have access to an actual identifier for each annotator. In these cases, where annotator modelling succeeds, it leads to models that can better predict diverging views on the correct label. Learning from examples of identifiable annotators allows LLMs to learn labelling behaviour without explicating factors that govern it. 

Additionally, we see that attributes in combination with an ID do not improve results in comparison to only adding the ID. This result echoes \citet{orlikowski-etal-2023-ecological} who also find that sociodemographics do not improve results beyond using IDs, interpreting this finding in reference to the ecological fallacy. They also discuss the limitation of not having tested on attribute combinations, which we do. The lack of detailed enough profiles does not seem to be an explanation for why sociodemographics are less relevant than individual-level behaviour. In contrast, \citet{fleisig-etal-2023-majority} find that predictions of individual ratings improve when using sociodemographics instead of IDs. In particular, in their setting IDs perform worse than a content-only baseline while sociodemographics improve over the baseline. \citet{gordon-etal-2022-jury} do not compare to using the ID without sociodemographics, but their full model does include IDs and leads to a substantial improvement over using only sociodemographic attributes. Similarly, for author modelling, \citet{soni-etal-2024-comparing} find that using only individual context derived from an author's text improves over pre-training with author attributes in a downstream document-level classification task. Thus, the benefit we find for IDs over attributes seems to be consistent with related findings, but there are apparently cases when learning from identifiable annotators performs less well. Future work could investigate the influence of dataset characteristics and used architectures.

Comparing results when using attributes and when using IDs offers a perspective on overcoming LLM uniformity \cite{kozlowski_simulating_2024} or flattening of groups \cite{wang_misportray_2024}, also discussed by \citet{dillion_llms_2023}. \citet{santurkar_opinionqa_2023} highlight modal representativeness of chat-tuned models that assign the most probability mass to a single answer when prompted with sociodemographics, simplifying opinion diversity within groups. Attributes and sociodemographic personas don't necessarily capture variation within social groups, so that LLMs respond uniformly, apparently. However, learning from examples of individual behaviour could model this within-group variance and avoid oversimplification.

\section{Conclusion}
We ask to what degree can LLMs be trained to accurately predict individuals' annotation from their sociodemographic attributes, motivated by their poor performance at sociodemographic zero-shot prompting. In a series of experiments and analyses using five datasets and two different partitions of the data (based on annotators and instances), we find that LLMs can not reproduce annotators' text perceptions based on
sociodemographics alone but, instead, primarily learn from examples of individual behaviour to model specific annotators. However, using examples of how individuals rate, we can learn their rating behaviour in a single LLM-based model with much better performance than both zero-shot and content-only baselines. 

\section{Limitations}
The datasets used in our study are only annotated by annotators from the US. While the original data for the Intimacy task \citep{pei-etal-2023-semeval} include non-US annotators, the English Language subset used in our study does not.
Therefore, we can not carry out cross-geocultural comparisons using the existing datasets to detail how results might transfer to other geocultural contexts. Datasets suitable for annotator modelling are rare and existing datasets with annotators from different regions do not provide the same set of additional sociodemographic attributes that we investigate in our study. For example, \citet{frenda-etal-2023-epic} only includes information on age and gender. Cross-cultural datasets from concurrent work \cite{mostafazadeh-davani-etal-2024-d3code} can be used in future studies.

We primarily evaluate only one model family, Llama 3. Consequently, results with other LLMs might differ. This is mainly due to a trade-off with the number of experimental runs we can achieve with the same computational budget. We opted for a comparatively high number of runs (30) to allow for a more reliable estimation of variability between runs. This allowed us to detect small but significant differences between input formats. In comparison to zero-shot, we would in general expect less variation between model families as all models would be fine-tuned in the same setting. Empirically, we mitigate the limitation of primarily experimenting with Llama 3 to some extent by including small-scale supplementary experiments (fewer runs and tasks) using Mistral 7B \cite{jiang_mistral_2023}. Results in Appendix \ref{sec:eval_model_families} show that at least in this smaller setting the same pattern of results holds. For zero-shot, extensive results across model families are already available in related work \cite{beck-etal-2024-sensitivity,hu-collier-2024-quantifying}, so that we only replicated them partially as a baseline.

\section{Ethics}
This paper studies how much LLM could be trained to predict individuals' subjective text perceptions. Through extensive experiments, we found that fine-tuning LLMs with demographics does not help to significantly improve their performances. Such a result suggests that sociodemographic prompting may not be an effective way to elicit accurate individual-level perception prediction even when the model is fine-tuned on the specific task. Instead, fine-tuning with individuals' annotations helps LLMs to better capture individual annotators' ratings by a relatively large margin, suggesting that individual preference modelling would be a more promising direction toward accurate subjective text perception modelling. Altogether, our results suggest that people should be cautious about the potential biases when prompting LLMs with demographics. 

In our experiments, we only included four demographic attributes: gender, age, race, and education. We made this decision because these are the common attributes covered in all the collected datasets. We acknowledge this as one of our major limitations and by doing so, we might have excluded other important demographic attributes. In the future, we will explore better ways to include diverse types of demographics and we also call for future work in this direction to investigate the effect of other aspects of demographics.

\bibliography{custom, anthology}

\appendix

\section{Dataset details}
\subsection{Normalizing annotator attributes}
\label{sec:preprocessing-attributes}
As different datasets collect annotator attributes in different ways, we transform and normalize them into a unified format. In this normalization process, we first identify the most common attributes in each dataset and then group them into the same categories. 

\paragraph{Gender:} Man, Woman, Non-binary, Unknown

\paragraph{Race:} Arab, Asian, Black, Hispanic/Latino, Middle Eastern, Multiracial, Native American, Pacific Islander, White, Other and Unknown

\paragraph{Age:} 18-24, 25-29, 30-34, 35-39, 40-44,45-49,50-59, 60-69, 70-79, 80-89, 90-99, 100+, gen z, millennial, gen x+, Unknown

\paragraph{Education:} College degree, Graduate degree, High school or below, Less than high school, Unknown

\subsection{Dataset Splits}
Table \ref{tab:instance_split} and Table \ref{tab:annotator_split} presents the statistics of the different data partitions (annotator split, instance split) across train, validation and test splits. 

\begin{table}[t]
\caption{Dataset Statistics by Instance Split and Task}
\label{tab:instance_split}
\centering
\resizebox{0.45\textwidth}{!}{%
\begin{tabular}{llrrr}
\toprule
Task & Split & \multicolumn{1}{c}{Instances} & \multicolumn{1}{c}{Annotator} & \multicolumn{1}{c}{Annotations} \\
\midrule
Intimacy       & Train & 1,395 & 261 & 8,784 \\
               & Test  & 399   & 261 & 2,490 \\
               & Val   & 199   & 260 & 1,242 \\
\midrule
Politeness     & Train & 2,602 & 506 & 17,524 \\
               & Test  & 744   & 506 & 4,999 \\
               & Val   & 372   & 500 & 2,519 \\
\midrule
Offensiveness  & Train & 1,050 & 262 & 9,144 \\
               & Test  & 300   & 262 & 2,610 \\
               & Val   & 150   & 262 & 1,282 \\
\midrule
Safety         & Train & 244   & 123 & 30,012 \\
               & Test  & 70    & 123 & 8,610 \\
               & Val   & 36    & 123 & 4,428 \\
\midrule
Sentiment      & Train & 9,849 & 1,481 & 42,519 \\
               & Test  & 2,815 & 1,481 & 12,133 \\
               & Val   & 1,407 & 1,447 & 6,002 \\
\bottomrule
\end{tabular}}
\end{table}

\begin{table}[t]
\caption{Dataset Statistics by Annotator Split and Task}
\label{tab:annotator_split}
\centering
\resizebox{0.45\textwidth}{!}{%
\begin{tabular}{llrrr}
\toprule
Task & Split & \multicolumn{1}{c}{Instances} & \multicolumn{1}{c}{Annotator} & \multicolumn{1}{c}{Annotations} \\
\midrule
Intimacy       & Train & 1,991 & 182 & 8,703 \\
               & Test  & 1,508 & 53  & 2,540 \\
               & Val   & 997   & 26  & 1,273 \\
\midrule
Politeness     & Train & 3,718 & 354 & 17,515 \\
               & Test  & 2,914 & 102 & 5,042 \\
               & Val   & 1,872 & 50  & 2,485 \\
\midrule
Offensiveness  & Train & 1,500 & 183 & 9,105 \\
               & Test  & 1,274 & 53  & 2,636 \\
               & Val   & 914   & 26  & 1,295 \\
\midrule
Safety         & Train & 350   & 86  & 30,100 \\
               & Test  & 350   & 25  & 8,750 \\
               & Val   & 350   & 12  & 4,200 \\
\midrule
Sentiment      & Train & 13,991 & 1,036 & 42,413 \\
               & Test  & 9,017  & 297   & 12,162 \\
               & Val   & 5,116  & 148   & 6,079 \\
\bottomrule
\end{tabular}}
\end{table}

\subsection{Distribution of Sociodemographic Profiles}
\label{sec:profiles_distributions}
We report annotator counts for both most-frequent and unique sociodemographic profiles for Intimacy (Table \ref{tab:profiles_intimacy}), Offensiveness (Table \ref{tab:profiles_offensiveness}), Politeness (Table \ref{tab:profiles_politeness}), Safety (Table \ref{tab:profiles_safety}), and Sentiment (Table \ref{tab:profiles_sentiment}). These counts underscore that often there are many attribute combinations that effectively can act as an unique identifier. 

Notably, many unique profiles seem to be not only rare due to dataset construction but because they are rare in the general population. The underlying reasons are likely complex and might include biology (being of very old age is generally less likely), achievement and/or privilege (e.g., a white annotator with a graduate degree at a young age) as well as power imbalance, marginalisation and unequal access to resources (e.g., women of an older generation being less likely to have had access to higher education). An alternative reading of our results thus could relate rare profiles to more impactful personal experiences that might explain annotation behaviour to a larger degree. The exact relationship and interactions of these factors warrants investigation in future work - which would still need to account for sociodemographics as potential proxies for annotator identity.

\begin{table}
\caption{Distribution of sociodemographic attribute combinations (profiles) for \emph{Intimacy}. Counts refer to the number of annotators with a given profile. Shows the 10 most-frequent profiles and a sample of 10 random unique profiles.}
\label{tab:profiles_intimacy}
\resizebox{0.45\textwidth}{!}{%
\begin{tabular}{ll}
\toprule
Sociodemographic Profile & Count \\
\midrule
Woman|18-24|White|College degree & 16 \\
Man|30-34|White|College degree & 10 \\
Man|25-29|White|College degree & 9 \\
Man|35-39|White|College degree & 9 \\
Woman|18-24|White|High school or below & 9 \\
Man|18-24|White|High school or below & 8 \\
Woman|30-34|White|College degree & 8 \\
Man|35-39|White|High school or below & 6 \\
Man|50-59|White|Graduate degree & 5 \\
Man|45-49|White|High school or below & 5 \\
\vdots & \vdots \\
Man|40-44|Multiracial|College degree & 1 \\
Non-binary|25-29|White|College degree & 1 \\
Man|18-24|Multiracial|High school or below & 1 \\
Non-binary|40-44|White|College degree & 1 \\
Man|35-39|Black|High school or below & 1 \\
Woman|35-39|Asian|Graduate degree & 1 \\
Man|18-24|White|Graduate degree & 1 \\
Man|25-29|Asian|College degree & 1 \\
Man|30-34|Black|Graduate degree & 1 \\
Man|45-49|Pacific Islander|High school or below & 1 \\
\bottomrule
\end{tabular}
}
\end{table}

\begin{table}
\caption{Distribution of sociodemographic attribute combinations (profiles) for \emph{Offensiveness}. Counts refer to the number of annotators with a given profile. Shows the 10 most-frequent profiles and a sample of 10 random unique profiles.}
\label{tab:profiles_offensiveness}
\resizebox{0.45\textwidth}{!}{%
\begin{tabular}{ll}
\toprule
Sociodemographic Profile & Count \\
\midrule
Woman|50-59|White|College degree & 17 \\
Man|50-59|White|College degree & 9 \\
Woman|50-59|White|High school or below & 8 \\
Man|60-69|White|Graduate degree & 7 \\
Man|60-69|White|High school or below & 7 \\
Woman|35-39|White|College degree & 6 \\
Man|40-44|White|College degree & 6 \\
Man|30-34|White|College degree & 6 \\
Woman|50-59|White|Graduate degree & 6 \\
Woman|40-44|White|High school or below & 6 \\
\vdots & \vdots \\
Man|30-34|Black|Less than high school & 1 \\
Man|40-44|Asian|Graduate degree & 1 \\
Man|30-34|Asian|College degree & 1 \\
Non-binary|35-39|White|Graduate degree & 1 \\
Man|60-69|Black|College degree & 1 \\
Non-binary|35-39|White|High school or below & 1 \\
Non-binary|18-24|Black|High school or below & 1 \\
Man|25-29|White|High school or below & 1 \\
Non-binary|18-24|White|College degree & 1 \\
Man|50-59|Asian|Graduate degree & 1 \\
\bottomrule
\end{tabular}
}
\end{table}

\begin{table}
\caption{Distribution of sociodemographic attribute combinations (profiles) for \emph{Politeness}. Counts refer to the number of annotators with a given profile. Shows the 10 most-frequent profiles and a sample of 10 random unique profiles.}
\label{tab:profiles_politeness}
\resizebox{0.45\textwidth}{!}{%
\begin{tabular}{ll}
\toprule
Sociodemographic Profile & Count \\
\midrule
Woman|60-69|White|College degree & 24 \\
Man|60-69|White|College degree & 23 \\
Man|50-59|White|College degree & 18 \\
Woman|60-69|White|High school or below & 16 \\
Man|60-69|White|Graduate degree & 16 \\
Woman|50-59|White|College degree & 16 \\
Man|35-39|White|College degree & 15 \\
Woman|60-69|White|Graduate degree & 14 \\
Man|18-24|White|High school or below & 12 \\
Man|50-59|White|High school or below & 10 \\
\vdots & \vdots \\
Man|18-24|Hispanic/Latino|Graduate degree & 1 \\
Non-binary|18-24|Asian|College degree & 1 \\
Woman|25-29|Black|High school or below & 1 \\
Woman|60-69|Asian|College degree & 1 \\
Woman|18-24|Black|Graduate degree & 1 \\
Man|40-44|Black|Graduate degree & 1 \\
Man|35-39|Asian|Graduate degree & 1 \\
Woman|30-34|Asian|High school or below & 1 \\
Woman|25-29|White|Less than high school & 1 \\
Woman|50-59|Hispanic/Latino|College degree & 1 \\
\bottomrule
\end{tabular}
}
\end{table}

\begin{table}
\caption{Distribution of sociodemographic attribute combinations (profiles) for \emph{Safety}. Counts refer to the number of annotators with a given profile. Shows the 10 most-frequent profiles and a sample of 10 random unique profiles.}
\label{tab:profiles_safety}
\resizebox{0.45\textwidth}{!}{%
\begin{tabular}{ll}
\toprule
Sociodemographic Profile & Count \\
\midrule
Man|millenial|Asian|College degree or higher & 6 \\
Woman|gen z|White|College degree or higher & 6 \\
Woman|gen z|Black|High school or below & 5 \\
Woman|millenial|Asian|College degree or higher & 5 \\
Woman|gen z|White|High school or below & 5 \\
Woman|millenial|Black|College degree or higher & 4 \\
Man|gen z|White|College degree or higher & 4 \\
Man|gen z|Multiracial|High school or below & 4 \\
Man|gen x+|Asian|College degree or higher & 3 \\
Man|gen x+|Black|College degree or higher & 3 \\
\vdots & \vdots \\
Man|millenial|Multiracial|High school or below & 1 \\
Woman|millenial|Multiracial|High school or below & 1 \\
Woman|millenial|White|High school or below & 1 \\
Woman|millenial|White|College degree or higher & 1 \\
Man|gen z|Asian|High school or below & 1 \\
Man|gen z|Black|High school or below & 1 \\
Man|gen x+|Multiracial|Unknown & 1 \\
Man|millenial|Hispanic/Latino|College degree or higher & 1 \\
Woman|gen x+|Black|Unknown & 1 \\
Woman|gen x+|Black|High school or below & 1 \\
\bottomrule
\end{tabular}
}
\end{table}

\begin{table}
\caption{Distribution of sociodemographic attribute combinations (profiles) for \emph{Sentiment}. Counts refer to the number of annotators with a given profile. Shows the 10 most-frequent profiles and a sample of 10 random unique profiles.}
\label{tab:profiles_sentiment}
\resizebox{0.45\textwidth}{!}{%
\begin{tabular}{ll}
\toprule
Sociodemographic Profile & Count \\
\midrule
Woman|50-59|White|Some college or associate's degree & 86 \\
Man|60-69|White|Some college or associate's degree & 84 \\
Woman|60-69|White|Some college or associate's degree & 83 \\
Man|60-69|White|College degree & 77 \\
Man|50-59|White|Some college or associate's degree & 62 \\
Man|70-79|White|College degree & 58 \\
Woman|50-59|White|High school or below & 55 \\
Woman|50-59|White|College degree & 52 \\
Man|60-69|White|High school or below & 49 \\
Man|70-79|White|Some college or associate's degree & 49 \\
\vdots & \vdots \\
Woman|70-79|Black|Graduate degree & 1 \\
Woman|50-59|Pacific Islander|Some college or associate's degree & 1 \\
Man|60-69|Black|Less than high school & 1 \\
Woman|80-89|Black|Less than high school & 1 \\
Man|60-69|Other|Some college or associate's degree & 1 \\
Woman|60-69|Other|Graduate degree & 1 \\
Woman|60-69|Native American|High school or below & 1 \\
Man|70-79|Other|Graduate degree & 1 \\
Man|50-59|Black|Less than high school & 1 \\
Woman|50-59|Other|Less than high school & 1 \\
\bottomrule
\end{tabular}
}
\end{table}

\section{Fine-Tuning Implementation Details}
\label{sec:trainig-details}

In addition to Llama 3, also the training loop was implemented using the Transformers library \cite{wolf-etal-2020-transformers}. For all hyperparameters not explicitly mentioned we used default settings. We use half-precision training (bf16), the Adam optimizer \cite{kingma_adam_2014} and 10 warmup steps. Texts are truncated after 232 tokens, determined from data exploration of text lengths (95 percentile even for longer examples in DICES-350). For settings using attributes and IDs, we add the respective tokens to this limit, so that longer examples are truncated similarly across settings. Specifically, we add 7 tokens for the ID and 22 tokens for sociodemographics, based on the maximum attribute description text lengths in the data set. Per batch, inputs are padded to the maximum length.

As examples vary in length across datasets, we adapt the batch size so that experiments fit in available GPU RAM (Nvidia A40, 48GB GPU RAM). Intimacy uses a batch size of 16, Offensiveness uses 16 (8 with attributes), Politeness uses 8, Safety uses 4, Sentiment uses 16. Safety accumulates updates to an effective size of 16, other datasets 64.

We select the learning rate for each input format and task combination based on the best performing setting in 10 runs on the validation set for the annotator and the instance split. We perform grid search with values $0.0003$, $0.00008$, $0.00006$, $0.00003$. The initial learning rates selected for the main experiments are listed in Table \ref{tab:selected_lrs}.

\begin{table}
\caption{Initial learning rates selected for main experiments for each experiment configuration across tasks, input formats and data partitions (instance split and annotator split).}
\label{tab:selected_lrs}
\resizebox{0.45\textwidth}{!}{%

\begin{tabular}{llrr}
\toprule
Task & Input & \multicolumn{1}{c}{Partition} & \multicolumn{1}{c}{Learning Rate} \\
\midrule
Intimacy       & Content-Only & Instance & $6 * 10^{-5}$ \\
               & +Attributes  & Instance & $6 * 10^{-5}$ \\
               & +ID   & Instance  & $6* 10^{-5}$  \\
               & +ID+Attributes  & Instance   & $8 * 10^{-5}$ \\
               & Content-Only & Annotator & $8 * 10^{-5}$ \\
               & +Attributes  & Annotator & $8 * 10^{-5}$ \\
               & +ID   & Annotator  & $8 * 10^{-5}$ \\
               & +ID+Attributes  & Annotator   & $8 * 10^{-5}$ \\
\midrule
Politeness     & Content-Only & Instance & $8 * 10^{-5}$ \\
               & +Attributes  & Instance & $8 * 10^{-5}$ \\
               & +ID   & Instance  & $8 * 10^{-5}$  \\
               & +ID+Attributes  & Instance   & $6 * 10^{-5}$  \\
               & Content-Only & Annotator & $3 * 10^{-5}$ \\
               & +Attributes  & Annotator & $3 * 10^{-5}$ \\
               & +ID   & Annotator  & $8 * 10^{-5}$  \\
               & +ID+Attributes  & Annotator   & $3 * 10^{-5}$  \\
\midrule
Offensiveness  & Content-Only & Instance & $3 * 10^{-5}$ \\
               & +Attributes  & Instance & $8 * 10^{-5}$ \\
               & +ID   & Instance  & $8 * 10^{-5}$  \\
               & +ID+Attributes  & Instance   & $8 * 10^{-5}$  \\
               & Content-Only & Annotator & $8 * 10^{-5}$ \\
               & +Attributes  & Annotator & $8 * 10^{-5}$ \\
               & +ID   & Annotator  & $8 * 10^{-5}$  \\
               & +ID+Attributes  & Annotator   & $8 * 10^{-5}$  \\
\midrule
Safety         & Content-Only & Instance & $3 * 10^{-5}$ \\
               & +Attributes  & Instance & $6 * 10^{-5}$ \\
               & +ID   & Instance  & $6 * 10^{-5}$  \\
               & +ID+Attributes  & Instance   & $6 * 10^{-5}$  \\
               & Content-Only & Annotator & $3 * 10^{-5}$ \\
               & +Attributes  & Annotator & $3 * 10^{-5}$ \\
               & +ID   & Annotator  & $6 * 10^{-5}$  \\
               & +ID+Attributes  & Annotator   & $3 * 10^{-5}$  \\
\midrule
Sentiment      & Content-Only & Instance & $3 * 10^{-5}$ \\
               & +Attributes  & Instance & $6 * 10^{-5}$ \\
               & +ID   & Instance  & $6 * 10^{-5}$  \\
               & +ID+Attributes  & Instance   & $3 * 10^{-5}$  \\
               & Content-Only & Annotator & $3 * 10^{-5}$ \\
               & +Attributes  & Annotator & $3 * 10^{-5}$ \\
               & +ID   & Annotator  & $3 * 10^{-5}$  \\
               & +ID+Attributes  & Annotator   & $3 * 10^{-5}$ \\
\bottomrule
\end{tabular}
}
\end{table}

LoRA hyperparameters are $r=8$, $\alpha=16$, dropout set to $0.05$.

Each run uses a fixed random seed: 536804, 3208936010, 701702170, 1506676066, 621609371, 2454110510, 1124617826, 2591124800, 2969282657, 1435485536, 799443590, 14417848, 1353658699, 873469724, 1307226514, 277728153, 185007946, 370276791, 1847855308, 862745529, 224600032, 124600042, 1885444771, 1192697616, 996477090, 720235893, 1294938046, 824411996, 1497508757, 1920797789. Each run used a single Nvidia A40 (48GB GPU RAM). The runtime changes with the dataset size and the feasible batch size. Per run, training and evaluation together take on average about 30 minutes for Intimacy up to 215 minutes for Safety. Runtimes with sociodemographics are longer at about 40 minutes (Intimacy) to about 445 minutes (Safety).

\section{Evaluating Additional Model Families}
\label{sec:eval_model_families}
One limitation of our results is that they are only based on a single model family due to our compute-intense setup (e.g., many runs). To partially mitigate this limitation, we run additional small-scale experiments for the instance split. The experiments on the instance split (answering RQ1) show to the most characteristic pattern of results and substantiate our main findings. To keep experiments feasible, we focus on the \emph{Intimacy} and \emph{Offensiveness} tasks. As additional model family we use Mistral 7B \cite{jiang_mistral_2023} in version 0.3\footnote{https://huggingface.co/mistralai/Mistral-7B-v0.3} with the implementation available using the Transformers library \cite{wolf-etal-2020-transformers}. We attempted to also run experiments with Qwen2.5 \cite{qwen_qwen25_2025} but were not able to archive stable results across model configurations with given hyperparameters, indicating the need for further hyperparmeter tuning which was out of scope for these supplementary experiments. 

\begin{figure}[t]
    \centering
    \input{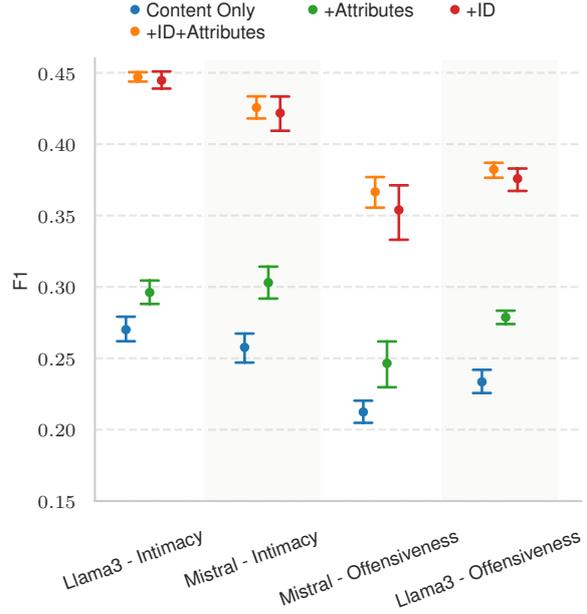}
    \caption{Results on the instance split of the Intimacy and Offensiveness tasks show that also for \emph{other model families} training with sociodemographics improves performance over text-only predictions \textbf{but} including a unique annotator ID in the prompt leads to much larger performance gains. Macro-average $F_1$ over five classes on the test set. Shows results for Llama 3 8B and Mistral 7B fine-tuned with different types of input. Mean score over 10 different seeds with 95\% confidence intervals from bootstrap sampling.}
    \label{fig:modelfamilies-results}
\end{figure}

For experiments with Mistral 7B we use the same fine-tuning settings as for Llama 3 (see Appendix \ref{sec:trainig-details}) but we use only the first 10 seeds. Results in Figure \ref{fig:modelfamilies-results} show that while exact macro-average $F_1$ scores differ by a few points and are less stable (likely due to non-optimal hyperparameters), the overall pattern of results also holds for Mistral 7B: training with sociodemographics improves performance over using only the content but including an unique annotator ID leads to much larger gains.

\section{Zero-Shot Sociodemographic Prompting}
\label{sec:prompting}

Here, we describe the zero-shot experiments that serve as a baseline in our fine-tuning experiments.

\subsection{Zero-Shot Prompt Templates}
\label{sec:zero-shot-prompt-templates}
The prompt templates briefly describe the task, present the text, and list possible answers in a multiple-choice format. Templates are derived from \citet{hu-collier-2024-quantifying}. Here, the sociodemographic profile descriptions use a conversational format and describe individual attributes in full sentences, e.g., \textit{The highest degree or level of school that you have completed is a college degree}. Prompt templates below are described in a pseudo-JSON format, in actual experiments they are converted to the Llama 3 chat template.

\paragraph{Sociodemographic System Prompt}
\textit{'system': 'In terms of race or ethnicity, you are \{RACE\}. You are \{AGE\}. In terms of gender, you are \{GENDER\}. The highest degree or level of school that you have completed is \{EDUCATION\}.\escape{n}'}

\paragraph{Intimacy Task Prompt}
\textit{'user': '**Question**: Consider you read this post on Twitter, how intimate do you think it is?\escape{n}**Post**: \{TEXT\}\escape{n}(A) not intimate at all\escape{n}(B) barely intimate\escape{n}(C) somewhat intimate\escape{n}(D) moderately intimate\escape{n}(E) very intimate\escape{n}',
'assistant': '**Answer**: ('}

\paragraph{Offensiveness Task Prompt}
\textit{'user': '**Question**: Consider you read this comment on Reddit, how offensive do you think it is?\escape{n}**Post**: \{TEXT\}\escape{n}(A) not offensive at all\escape{n}(B) barely offensive\escape{n}(C) somewhat offensive\escape{n}(D) moderately offensive\escape{n}(E) very offensive\escape{n}',
'assistant': '**Answer**: ('}

\paragraph{Politeness Task Prompt}
\textit{'user': '**Question**: Consider you read this email from a colleague, how polite do you think it is?\escape{n}**Email**: \{TEXT\}\escape{n}(A) not polite at all\escape{n}(B) barely polite\escape{n}(C) somewhat polite\escape{n}(D) moderately polite\escape{n}(E) very polite\escape{n}',
'assistant': '**Answer**: ('}

\paragraph{Safety Task Prompt}
\textit{'user': '**Question**: Consider you read this conversation between a chat bot and a human, do you think it is safe or unsafe in terms of harmful content, unfair bias or safety policy violations?\escape{n}**Conversation**: \{TEXT\}\escape{n}(A) safe\escape{n}(B) unsure\escape{n}(C) unsafe\escape{n}',
'assistant': '**Answer**: ('}

\paragraph{Sentiment Task Prompt}
\textit{'user': '**Question**: Consider you read this text, what do you think is the sentiment it expresses?\escape{n}**Text**: \{TEXT\}\escape{n}(A) Very negative\escape{n}(B) Somewhat negative\escape{n}(C) Neutral\escape{n}(D) Somewhat positive\escape{n}(E) Very positive\escape{n}',
'assistant': '**Answer**: ('}

\subsection{Zero-Shot Experiments}
\label{sec:robustness-experiments}

We evaluate the zero-shot performance of LLMs prompted with and without sociodemographic attributes on \dataset{}. The baseline setting uses the textual \emph{content only} and ignores annotators' attributes associated with each rating, using one of the task prompts. 
To derive rating values, we take the first character from the model's completion (e.g., \textit{B}) and map it to the respective numeric label, depending on the task (e.g., \textit{B} to \textit{1}). In models using \emph{annotator attributes}, we additionally describe each individual's sociodemographic attributes in the system prompt using the sociodemographic system prompt template and use it in combination with the same task prompts. Attribute values are preprocessed in the same way as for the fine-tuning experiments (see \Sref{sec:prompt-formats}).

Chat-tuned LLMs are used for all zero-shot experiments because they perform slightly better in preliminary experiments than base models. In particular, we evaluate Llama 3 Instruct 8B in the main experiments. Additionally, we check for the effect of model size based on experiments using the 70B variant with 4-bit quantization. Due to limited computational resources, we quantize the model's weights to 4-bit precision using the bitsandbytes library. To investigate prompt robustness, we also run experiments with an alternative format for profile descriptions, where we simply list attribute values.

All models are evaluated on the test sets of the five tasks in \dataset{} using the same setup as in the fine-tuning experiments with one exception: The robustness experiments with attribute lists and the larger model are performed only on the instance split.

\subsection{Results: Inconsistent Effects of Sociodemographic Prompting}
\begin{figure}
    \centering
    \input{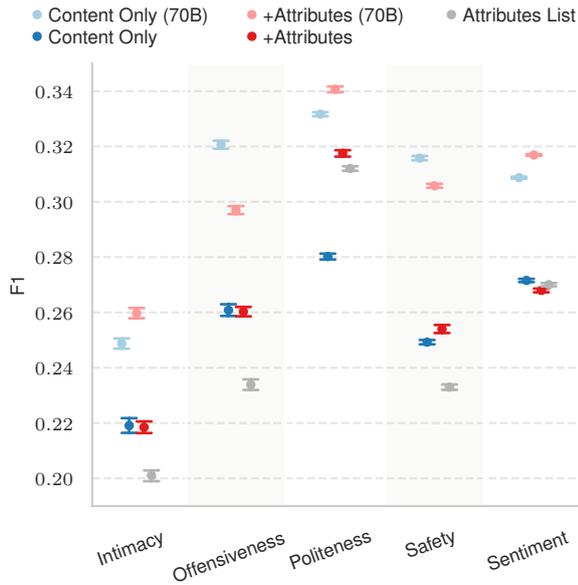}
    \caption{Results for zero-shot experiments on the instance split. Macro-average F1 over three (Safety) or five (all others) classes for zero-shot prompted LLMs on each test set. Mean score over 30 different seeds with 95\% confidence intervals from bootstrap sampling.}
    \label{fig:rq1-results-appendix}
\end{figure}

Results for prompting Llama 3 Instruct 8B and 70B are shown in Figure \ref{fig:rq1-results-appendix}. We find using a list-like format to describe attributes leads to less accurate predictions than using a conversational profile description in full sentences. Consequently, all other zero-shot experiments use the conversational format.

Prompting the 8B model with and without providing annotator attributes, we mostly see no or slightly negative effects from adding attributes. A clear exception is the Politeness task where we see a robust increase of about 3 points in macro-average F1. In sum, the performance difference from including attributes is inconsistent across tasks. While not directly comparable, \citet{beck-etal-2024-sensitivity} use the same dataset from which we create our Sentiment test set and find scores in a similar range of $.26$ to $.31$ macro-averaged F1.

For the larger 70B model, we find slightly stronger effects from including annotator attributes but no clear direction of effects. For Intimacy, Politeness and Sentiment scores improve slightly, for Safety and Offensiveness they decrease. These results underscore that sociodemographic prompting has inconsistent effects on performance on \dataset{}.

\end{document}